\documentclass[10pt,conference]{IEEEtran}
\usepackage{amsmath,amssymb}
\usepackage{microtype}
\usepackage[boxed,linesnumbered]{algorithm2e}
\usepackage{mathtools,graphicx,epsfig}
\usepackage[caption=false,font=footnotesize]{subfig}

\ifCLASSINFOpdf
\else
\fi

\begin{document}

\title{Fast and High-Quality Bilateral Filtering Using Gauss-Chebyshev Approximation}

\author{
  \IEEEauthorblockN{Sanjay Ghosh \quad \quad Kunal N. Chaudhury\thanks{This work was partially supported by the Startup Grant awarded by the Indian Institute of Science.}}
  \IEEEauthorblockA{Department of Electrical Engineering,
		    Indian Institute of Science, Bangalore, India\\	
		    e-mail: \{sghosh,kunal\}@ee.iisc.ernet.in}
}

\IEEEoverridecommandlockouts
\IEEEpubid{\makebox[\columnwidth]{978-1-5090-1746-1/16/\$31.00~\copyright~2016~IEEE \hfill}
\hspace{\columnsep}\makebox[\columnwidth]{ }}

\maketitle

\begin{abstract}
The bilateral filter is an edge-preserving smoother that has diverse applications in image processing, computer vision, computer graphics, and computational photography.
The filter uses a spatial kernel along with a range kernel to perform edge-preserving smoothing.
In this paper, we consider the Gaussian bilateral filter where both the kernels are Gaussian.
A direct implementation of the Gaussian bilateral filter requires $O(\sigma_s^2)$ operations per pixel, where $\sigma_s$ is the standard deviation of the spatial Gaussian. 
In fact, it is well-known that the direct implementation is slow in practice. 
We present an approximation of the Gaussian bilateral filter, whereby we can cut down the number of operations to $O(1)$ per pixel for any arbitrary $\sigma_s$, and yet achieve very high-quality filtering that is almost indistinguishable from the output of the original filter. We demonstrate that the proposed approximation is few orders faster in practice compared to the direct implementation. We also demonstrate that the approximation is competitive with existing fast algorithms in terms of speed and accuracy. 
\end{abstract}


\section{Introduction}
\label{sec:INTRO}

The bilateral filter was proposed by Tomasi and Maduchi \cite{Tomasi1998} as a non-linear extension of the classical Gaussian  filter. It is an instance of an edge-preserving filter that can simultaneously smooth homogenous regions and preserve sharp edges. We refer the reader to \cite{Book2009} for a detailed survey of the working of the filter and its various applications.

Before proceeding further, we set up the necessary notation and terminology. Consider an image $f : I \rightarrow \mathbb{R}$, where $I$ is some finite rectangular domain of $\mathbb{Z}^2$; we extend the image outside $I$ using symmetry, if required. Consider the Gaussian \textit{kernels} 
$g_{\sigma_s} : \mathbb{Z}^2 \rightarrow \mathbb{R}$ and $g_{\sigma_r} : \mathbb{R} \rightarrow \mathbb{R}$ given by
\begin{equation*}
 g_{\sigma_s}(i) = \exp \ \Big(- \frac{|| i ||^2}{2 \sigma_s^2} \Big) \ \text{ and } \ 
 g_{\sigma_r}(t) = \exp \Big(- \frac{t^2}{2 \sigma_r^2} \Big).
\end{equation*}
The former is called the spatial kernel and the latter is called the range kernel \cite{Tomasi1998}.
The output of the Gaussian bilateral filter is the image $\mathcal{B}[f]: I \rightarrow \mathbb{R}$ given by 
\begin{equation}
 \mathcal{B}[f](i) = \frac{\sum_{j\in \Omega} g_{\sigma_s}(j) \ g_{\sigma_r}( f(i-j) - f(i)) f(i-j)}{\sum_{j\in \Omega} g_{\sigma_s}(j) \  g_{\sigma_r}( f(i-j) - f(i))}.
 \label{BF}
\end{equation}
In practice, the domain of the spatial kernel $\Omega$ is set to be $\Omega=[-W,W] \times [-W,W]$, where $W=3\sigma_s$ \cite{Tomasi1998}. 

It is clear that the direct computation of \eqref{BF} requires $O(\sigma_s^2)$ operations per pixel. In general, the direct implementation  is slow for practical settings of $\sigma_s$ \cite{Book2009,Paris2006}. To overcome this, researchers have come up with several fast algorithms; e.g., see \cite{Paris2006,Porikli2008,Yang2009,Chaudhury2011,Chaudhury2013,Chaudhury2015}. We refer the interested reader to  \cite{Book2009,Chaudhury2015} for a survey of  fast algorithms. The fast algorithms in \cite{Porikli2008,Chaudhury2011,Chaudhury2013} are particularly relevant to the present work. The authors here approximate the Gaussian range kernel using polynomial and trigonometric functions respectively, and demonstrate how the resulting filter can be decomposed into a series of spatial Gaussian filtering. The degree (or order) of the spatial filtering is used to control the filtering accuracy; the larger the degree, the better is the accuracy. On the other hand, since the Gaussian filter can be implemented using $O(1)$ operations per pixel (e.g., using separability and recursion \cite{Deriche1993}), the overall approximation has $O(1)$ complexity with respect to $\sigma_s$ as a result. 
Here and henceforth, by filtering accuracy, we will refer to the quality of the filtering in relation to \eqref{BF} that can be obtained using a given approximation. 
More recently, a novel approach was proposed in \cite{Chaudhury2015}, where the author propose to approximate the range kernel using the so-called Gauss-polynomials. Similar to \cite{Porikli2008,Chaudhury2013}, the Gauss-polynomial approximation allows one to decompose the bilateral filter into a series of Gaussian filtering. In particular, it was demonstrated in \cite{Chaudhury2015} that the Gauss-polynomial approximation offers better accuracy compared to the polynomial approximation \cite{Porikli2008}, and is more amenable to hardware implementation in contrast to the trigonometric approximations \cite{Chaudhury2011,Chaudhury2013}. 

As will be demonstrated shortly, it turns out that the Gauss-polynomial approximation can result in low-accuracy filtering for images with lot of sharp edges, unless the degree of the polynomial is very high (the run-time, however, scales with the degree). We trace the source of this problem to the Taylor polynomials used in \cite{Chaudhury2015} to approximate the exponential function over an interval. In particular, while a Taylor polynomial of fixed degree provides good approximation near the origin, the approximation tends to degrade as one moves away from the origin. In other words, the  error incurred by the approximation is not evenly distributed over the interval. As is well-known, this problem can be fixed by replacing the Taylor polynomial with a Chebyshev polynomial of the same degree. Indeed, the Chebyshev polynomials have the property that they balance out the error over the whole interval \cite{EPPT1987}. We demonstrate that the filtering accuracy can indeed be increased (often substantially) by replacing the Taylor polynomial with a Chebyshev polynomial of the same degree. While the resulting algorithm structurally resembles the one in \cite{Chaudhury2015}, the filtering now is shown to be much more accurate for images with sharp edges. Importantly, the run-time of the proposed algorithm is almost identical to that of the algorithm in \cite{Chaudhury2015}, and we retain the advantage of using polynomials that is advantageous for hardware implementations.

The rest of the paper is organized as follows. We recall the Gauss-polynomial approximation proposed in \cite{Chaudhury2015} and the resulting fast algorithm in Section \ref{sec:FBF}. We next propose the Gauss-Chebyshev approximation in Section \ref{sec:GCA}. We present some simulation results obtained using the proposed fast algorithm in Section \ref{sec:SIM}, before concluding the paper.

\section{Fast Bilateral Filtering}
\label{sec:FBF}

The proposed fast algorithm is an extension of the algorithm in \cite{Chaudhury2015}. We now review the main ideas behind the fast algorithm in \cite{Chaudhury2015}. The author here proceeds by factoring the range kernel $g_{\sigma_r}(t-\tau)$ in \eqref{BF}, where $t = f(i-j)$ and $\tau = f(i)$, as follows:
\begin{equation}
\label{factor}
 g_{\sigma_r}(t-\tau) = \exp \Big(\!- \frac{\tau^2}{2 \sigma_r^2} \Big) \ \exp \Big(- \frac{t^2}{2 \sigma_r^2} \Big) \ \exp \left( \frac{\tau t}{ \sigma_r^2} \right).
\end{equation} 
The variables $t$ and $\tau$ take on values in some intensity range $[L, U]$. For example, $L=0$ and $U=255$ for a 8-bit grayscale image. The proposal in \cite{Chaudhury2015} is to use Taylor polynomials to approximate the third exponential term in \eqref{factor}; the advantage of doing so will be evident shortly. In particular, setting $x= \tau t/ \sigma_r^2$, the following Taylor approximation is considered
\begin{equation*}
\exp(x) \approx \sum_{n=0}^N \frac{x^n}{n!}.
\end{equation*}
The key observation in this context is that there is nothing special about Taylor polynomials, and that any arbitrary polynomial could be used. Of course, the advantage with Taylor polynomials is that the coefficients have a particularly simple expression. Continuing with our observation, we consider a general polynomial approximation
\begin{equation}
\label{poly_approx}
\exp(x) \approx \sum_{n=0}^N c_n x^n,
\end{equation}
where $N$ is the \textit{degree} of the polynomial. Substituting \eqref{poly_approx} in \eqref{factor}, we arrive at the following approximation of \eqref{factor}:
\begin{equation}
\label{kernel_approx}
\exp \Big(- \frac{\tau^2}{2 \sigma_r^2} \Big) \ \exp \Big(- \frac{t^2}{2 \sigma_r^2} \Big) \  \sum_{n=0}^N c_n  \left( \frac{\tau t}{ \sigma_r^2} \right)^n.
\end{equation}
Being the product of two Gaussians and a polynomial, \eqref{kernel_approx} was referred to as the \textit{Gauss-Polynomial} approximation in \cite{Chaudhury2015}.
We substitute \eqref{kernel_approx} into \eqref{BF}, and compute the numerator and the denominator of the resulting approximation. Recalling that $t = f(i-j)$ and $\tau = f(i)$, the numerator is given by
\begin{align}
\label{double_sum}
&\sum_{j\in \Omega} g_{\sigma_s}(j) f(i-j) \exp \left(- \frac{f(i)^2}{2 \sigma_r^2} \right) \nonumber  \\ 
& \exp \left(- \frac{f(i-j)^2}{2 \sigma_r^2} \right) \  \sum_{n=0}^N c_n  \left( \frac{f(i) f(i-j)}{ \sigma_r^2} \right)^n.
\end{align}
Next, for $n= 0,\ldots,N$, we construct the images $G_n : I \rightarrow \mathbb{R}$ and $F_n : I \rightarrow \mathbb{R}$ given by
\begin{equation}
\label{intImg}
 G_n(i) = \left(\frac{f(i)}{\sigma_r} \right)^n \ \text{and}  \ F_n(i) = \exp \left( \! - \frac{f(i)^2}{2 \sigma_r^2} \right) G_n(i).
\end{equation}
We then consider the output of the Gaussian filtering $\bar{F}_n : I \rightarrow \mathbb{R}$ of each $F_n$, given by
\begin{equation}
\label{GaussFilter}
 \bar{F}_n(i) = \sum_{j \in \Omega} g_{\sigma_s}(j) F_n(i-j).
 \end{equation}
By exchanging the order of the summations, and after some manipulation, we can write \eqref{double_sum} as $\exp\left(- f(i)^2/2 \sigma_r^2 \right) P(i)$, where
\begin{equation}
\label{P}
P(i) =  \sigma_r \sum_{n = 0}^{N} c_n G_n (i) \bar{F}_{n+1} (i). 
\end{equation}
In an identical manner, we can approximate the denominator using $\exp\left(- f(i)^2/2 \sigma_r^2 \right) Q(i)$, where
\begin{equation}
\label{Q}
Q(i) = \sum_{n = 0}^{N} c_n G_n (i) \bar{F}_n (i).
\end{equation}
In other words, the approximation of \eqref{BF} obtained using the kernel approximation in \eqref{kernel_approx} is given by
\begin{equation}
\label{approxBF}
\widehat{\mathcal{B}[f]}(i) = \frac{P(i)}{Q(i)}. 
\end{equation}
As remarked in \cite{Chaudhury2015}, using the range approximation in \eqref{kernel_approx}, we have effectively transferred the non-linearity of the bilateral filter to the intermediate images in \eqref{intImg}, which are obtained from the input image using simple pointwise transforms. The computational edge that we get from the above manipulation is that the filtering in \eqref{GaussFilter} can be computed using $O(1)$ operations per pixel  for any arbitrary $\sigma_s$ \cite{Deriche1993}. The overall cost of computing \eqref{approxBF} is therefore $O(1)$ per pixel with respect to $\sigma_s$. On the other hand, the complexity is $O(N)$ with respect to $N$, since we are required to compute the $N+2$ images $\bar{F}_0,\ldots,\bar{F}_{N+1}$ in \eqref{P} and \eqref{Q}. Needless to say, we can obtain better approximation (and hence better accuracy) by increasing $N$, but at the expense of the run-time.

\section{Gauss-Chebyshev Approximation}
\label{sec:GCA}

In this section, we present the so-called \textit{Gauss-Chebyshev} approximation in which we use the Chebyshev approximation of $\exp(x)$. This essentially amounts to computing the coefficients $(c_n)$  in \eqref{poly_approx}. The motivation behind this choice comes from the well-known fact in approximation theory that, for a fixed degree, the Chebyshev approximation outperforms the Taylor approximation \cite{EPPT1987,BJ2010}. 
In fact, Taylor polynomials can be fairly poor to approximate a function, except perhaps on very small intervals.
The ideal choice would have been the so-called \textit{minimax polynomial} that minimizes the maximum absolute error ($\ell_{\infty}$ error).
However, finding such a polynomial is computationally expensive, and hence it is a common practice to settle for a near-minimax approximation such as that obtained using Chebyshev polynomials \cite{EPPT1987,PFTV1990}. We now present some basic facts about Chebyshev polynomials before turning to the proposed Gauss-Chebyshev approximation.

\subsection{Chebyshev Approximation}
 
The Chebyshev polynomials $T_0(x), T_1(x), \ldots$ are defined over the interval $[-1,1]$. They are given by the formula \cite{EPPT1987}
\begin{equation}
\label{def}
 T_l(x) = \cos (l \arccos x), \qquad x\in [-1,1].
\end{equation}
The lowest-order polynomials are explicitly given by: $T_0(x) = 1, T_1(x) = x, T_2(x) = 2x^2 - 1$, and so on. The higher-order polynomials can be computed using the  recurrence
\begin{equation}
\label{recursion}
 T_{l+1}(x) = 2 x T_l(x) - T_{l-1}(x).
\end{equation}
This follows from definition \eqref{def}. Let $a_{l,n}$ be the coefficient of $x^n$ in $T_l(x)$, so that
\begin{equation}
\label{expansion}
T_l(x) = \sum_{n=0}^{l} a_{l,n} x^n.
\end{equation}
Notice that \eqref{recursion} induces a recursion on the coefficients $(a_{l,n})$. This can be used to compute $(a_{l,n})$ in an efficient and stable manner \cite{PFTV1990}. The coefficients can also be computed offline and stored in a look-up table. It follows from \eqref{def} that $T_l(x)$ has $l$ zeros over the interval $[-1, 1]$, given by
\begin{equation*}
 \xi_k = \text{cos} \Big[ \frac{\pi (2k - 1)}{2 l} \Big] \qquad (k = 1,2,\ldots, l).
\end{equation*}
The Chebyshev polynomials satisfy a type of ``discrete'' orthogonality relationship \cite{PFTV1990} in that
\begin{equation}
\label{ortho}
\small{
\sum_{k=1}^{l} T_i(\xi_k) T_j(\xi_k) = 
    \begin{dcases}
    0 \quad  \ & i \neq j, \\
    {l}/{2}  & i   =  j  \neq 0,\\
    l & i = j = 0.
  \end{dcases}}
\end{equation}

A given function $h(x), x \in [-1, 1]$ can be approximated in terms of Chebyshev polynomials using the linear expansion 
\begin{equation}
\label{Cheb_approx}
 h(x) \approx \sum_{l = 0}^{N} d_l T_l(x),
\end{equation}
where $(d_l)$ are the coefficients of the expansion. Note that the function on the right is a polynomial of degree $N$.
A particularly simple means of fixing the coefficients is to sample $h(x)$ at the zeros $\{\xi_k : k=1,\ldots,N+1\}$ of $T_{N+1}(x)$.
Then, by exploiting the orthogonality relationship in \eqref{ortho}, we get
\begin{equation}
\label{coeff}
 d_l =  \begin{dcases}
 \frac{1}{N+1} \sum_{k=1}^{N+1} \! h(\xi_k), \ \  & l = 0\\
 \frac{2}{N+1} \sum_{k=1}^{N+1} \! h(\xi_k) T_l(\xi_k),  \ \ & l \neq 0.
 \end{dcases}
\end{equation}
The approximation in \eqref{Cheb_approx} obtained using  \eqref{coeff} is very close to the minimax approximation of $h(x)$ \cite{PFTV1990}.

\begin{figure}
\centering
\subfloat[Pointwise error.]{\label{fig3}\includegraphics[width= 0.5\linewidth]{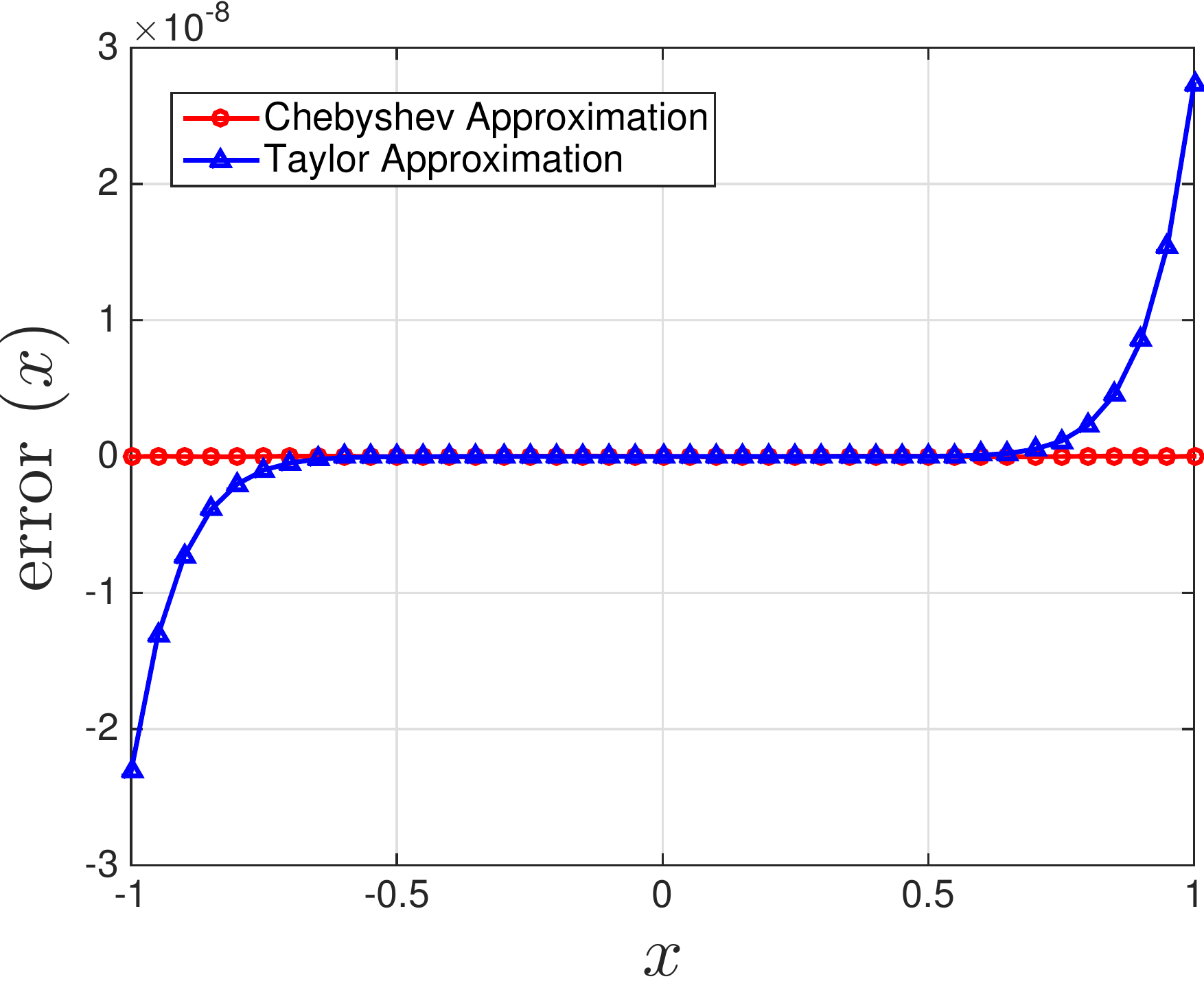}} 
\subfloat[$\ell_{\infty}$ error.]{\label{fig4}\includegraphics[width= 0.51\linewidth]{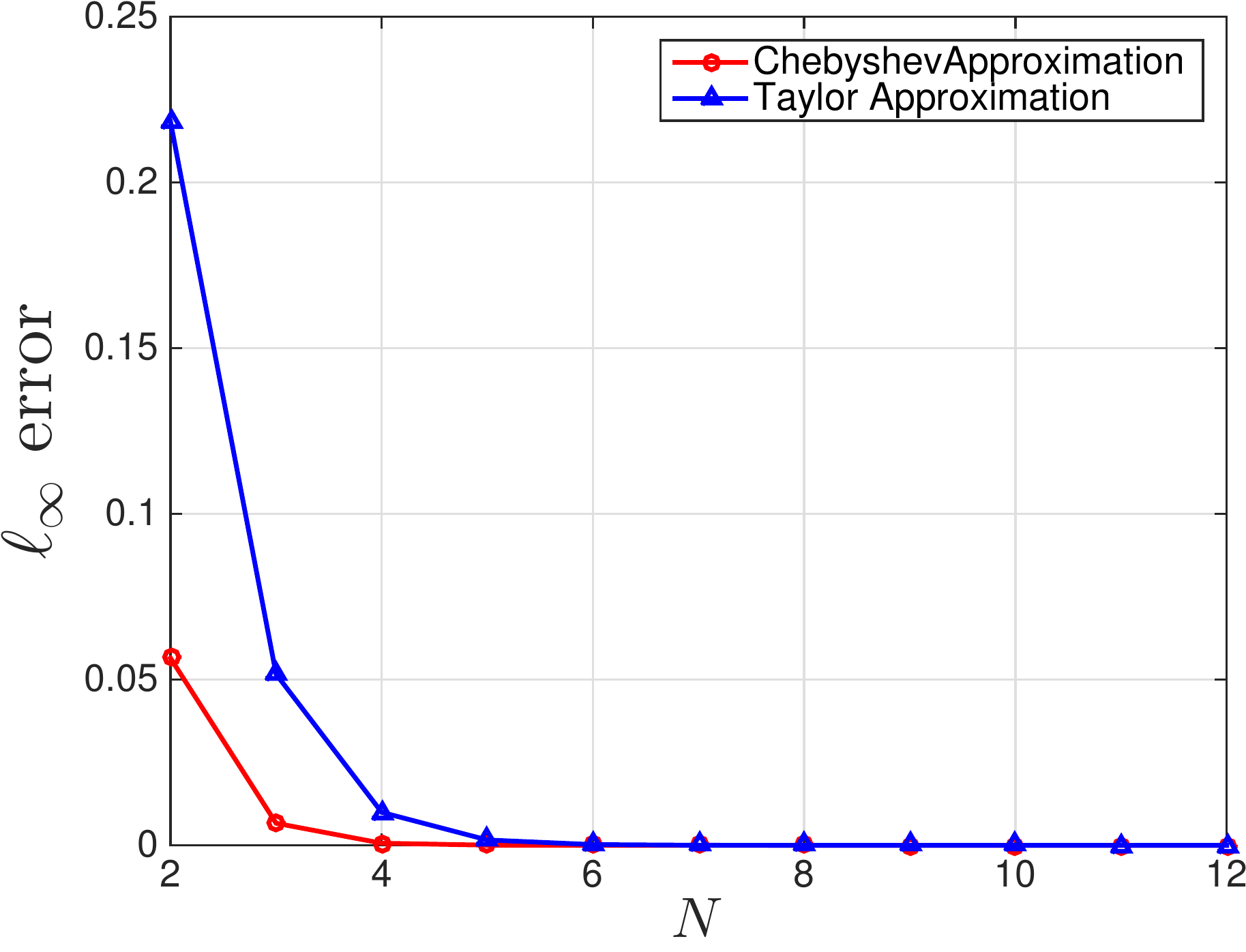}}\\
\caption{Comparison of the Chebyshev and Taylor approximations for $\exp(x)$ over the interval $[-1,1]$. Plot (a) shows the pointwise error when $N=10$. Plot (b) depicts the approximation error (measured using the $\ell_{\infty}$ norm) as a function of the approximation degree $N$.}
\label{comparison}
\end{figure}

\subsection{Kernel Approximation and Filtering}

The present idea is to approximate the exponential function (third factor) in \eqref{factor} using Chebyshev polynomials. 
In particular, we use the Chebyshev approximation in \eqref{Cheby} for the polynomial in \eqref{poly_approx}.
This choice is motivated using a numerical example in Figure \ref{comparison}, where we compare the approximations achieved using Taylor and Chebyshev polynomials of the same degree. .jpg
We see that the Taylor  polynomial tends to perform poorly as one moves away from the origin. On the other hand, the Chebyshev counterpart works well over the entire interval.

\begin{figure}
\centering
\subfloat[Error using Taylor.]{\label{fig3}\includegraphics[width=0.50 \linewidth]{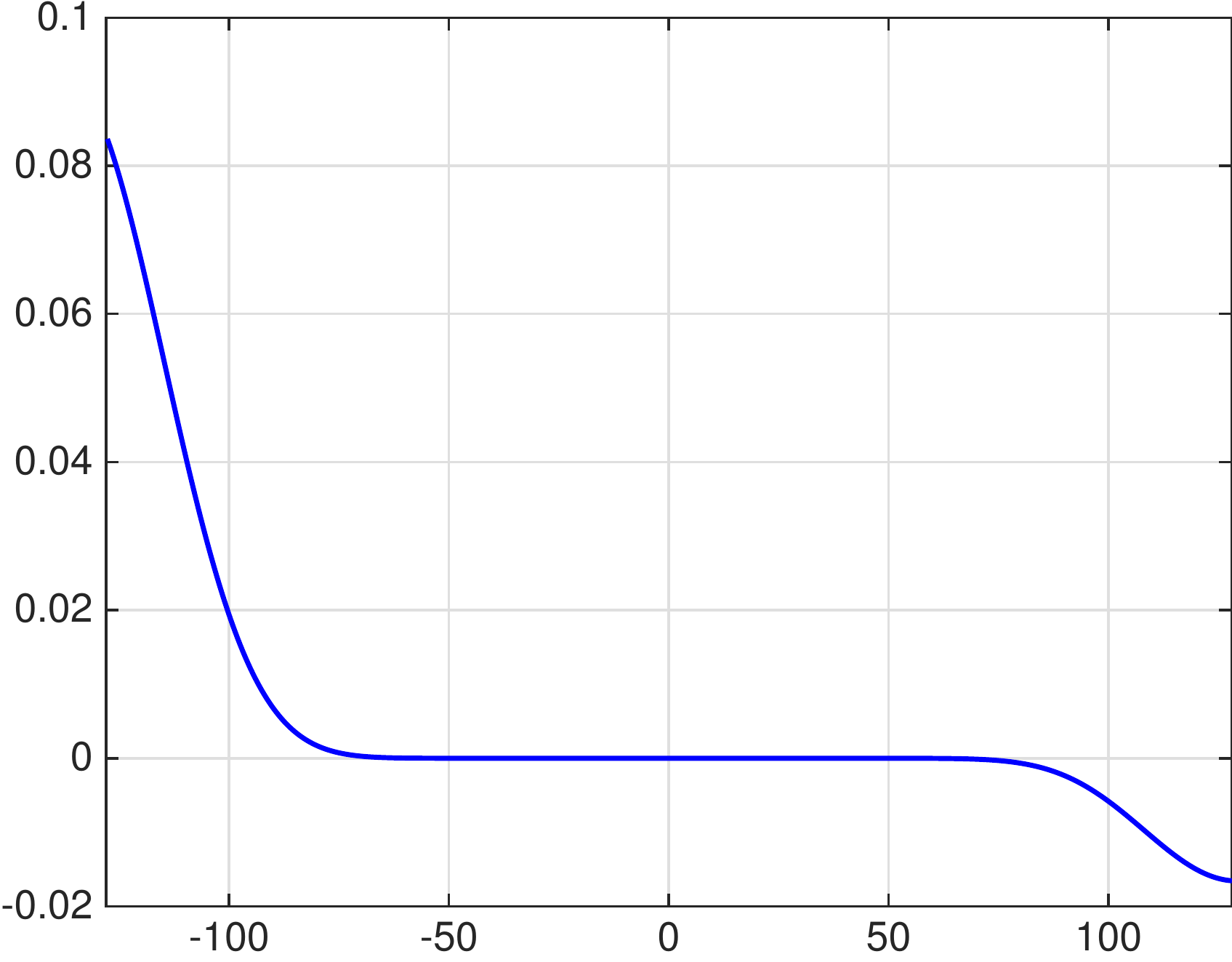}} 
\subfloat[Error using Chebyshev.]{\label{fig4}\includegraphics[width= 0.48 \linewidth]{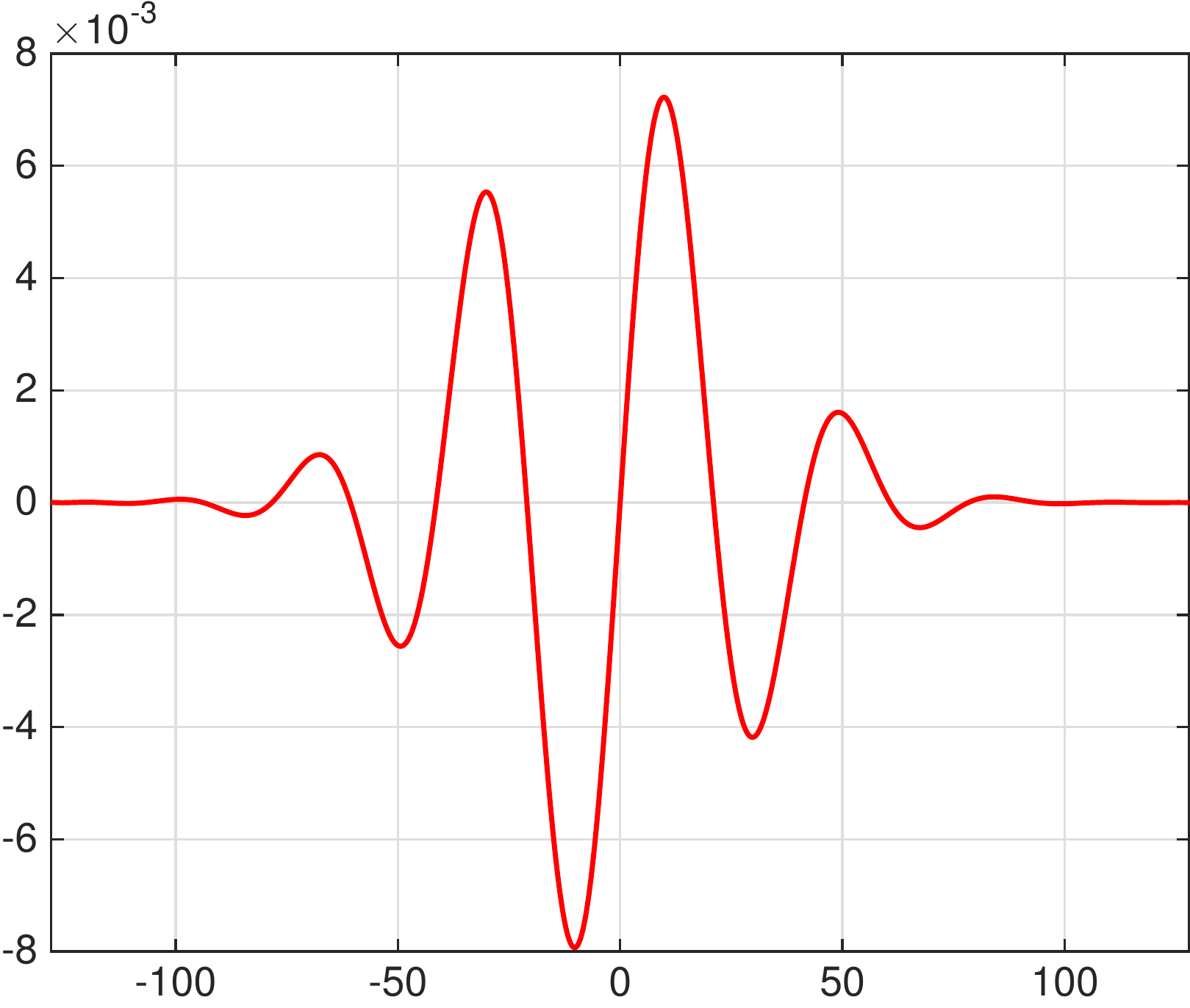}}
\caption{Pointwise error incurred in approximating $g_{30}(t -  100)$ using Taylor and Chebyshev polynomials of degree $N = 20$. The approximation interval is the centered dynamic range $[-t_c,t_c]$, where $t_c=127.5$ for an $8$-bit image.}
\label{fig:tau100}
\end{figure}

\begin{figure}
\centering
\subfloat[Taylor approximations.]{\label{fig3}\includegraphics[width=0.50 \linewidth]{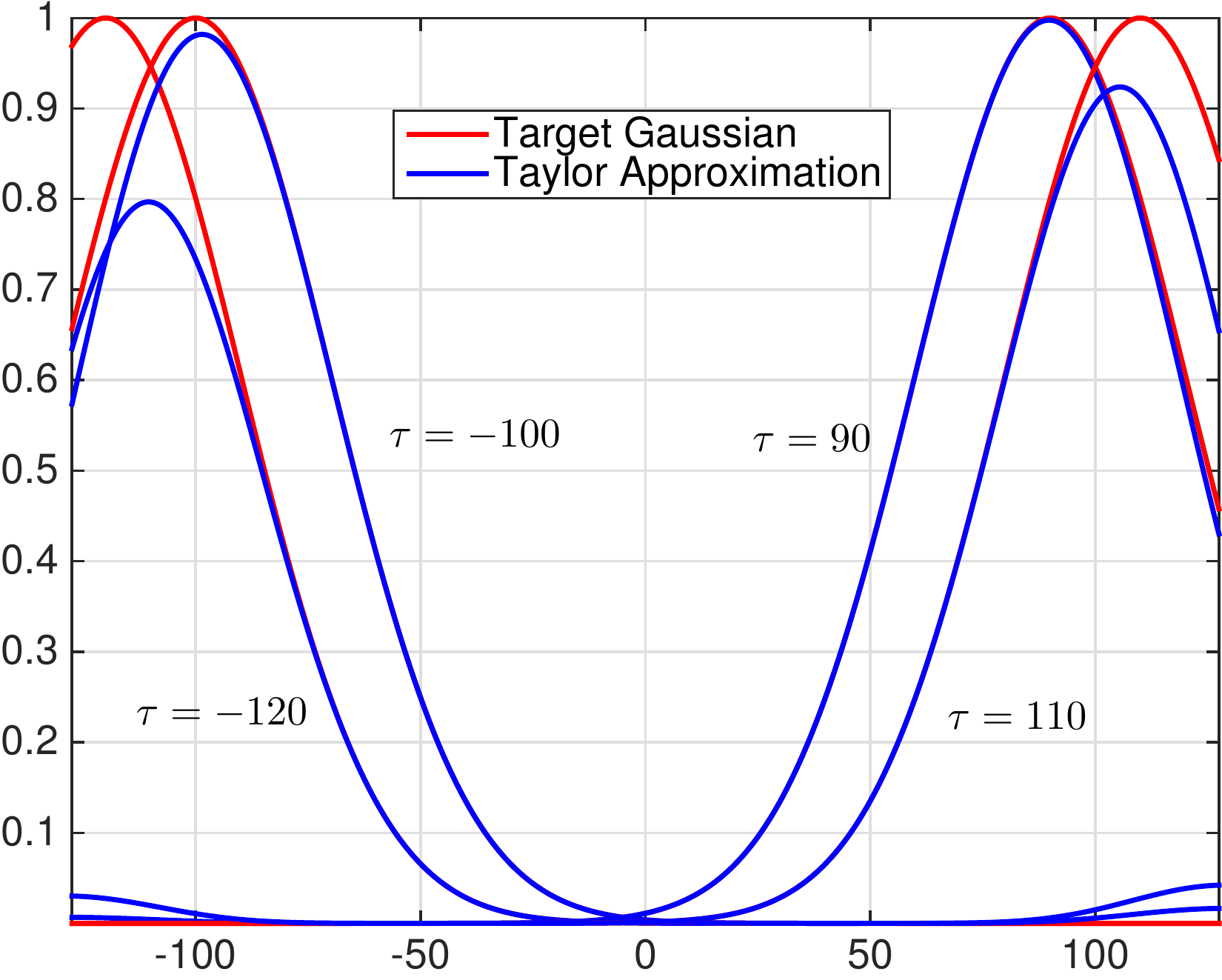}} 
\subfloat[Chebyshev approximations.]{\label{fig4}\includegraphics[width= 0.50 \linewidth]{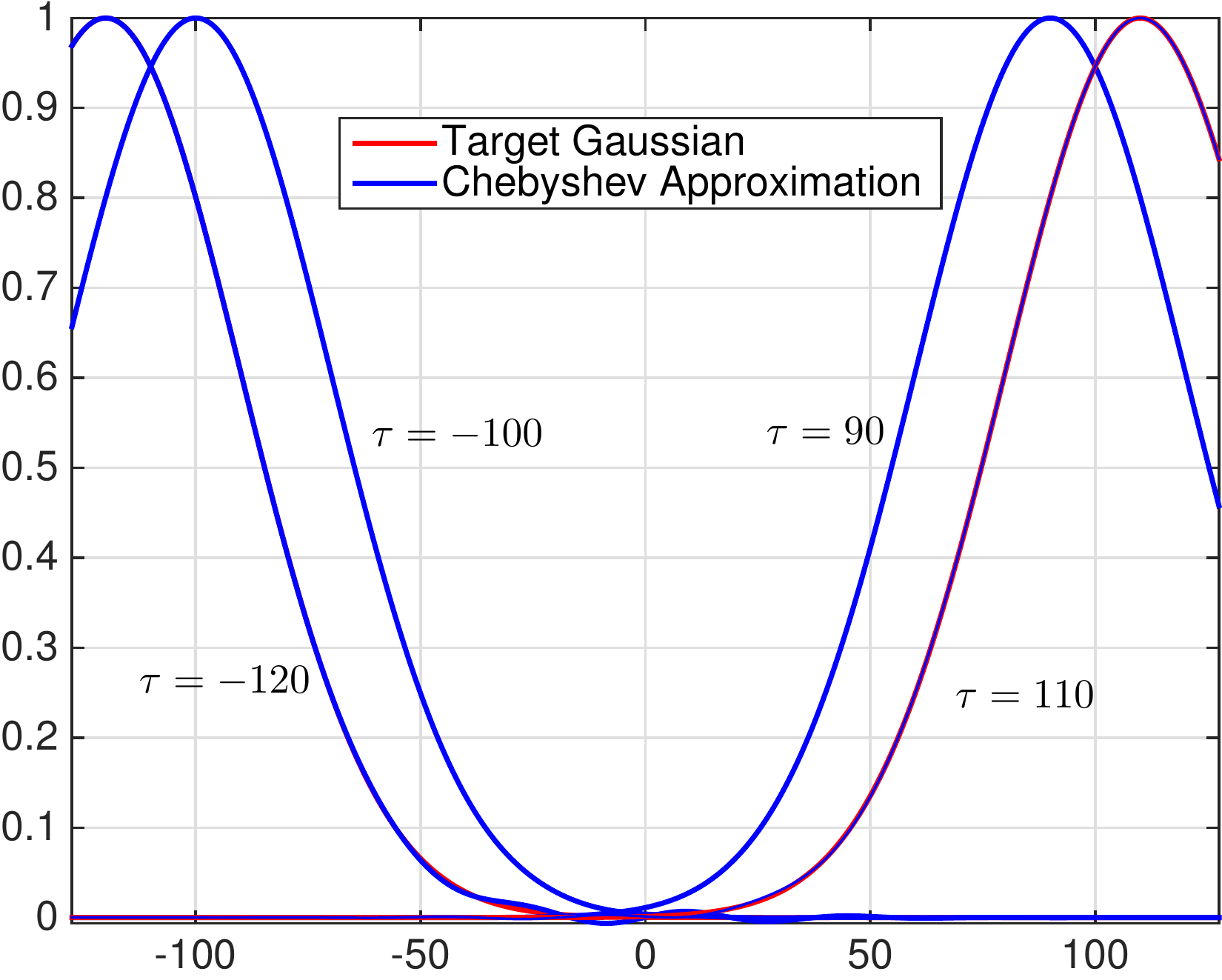}}
\caption{Approximation of $g_{30}(t - \tau)$ at $\tau \in \{-120, -100, 90, 110\}$ using Taylor and Chebyshev polynomials of degree $N = 20$.} 
\label{fig:vary_tau}
\end{figure}

There are, however, some details that we must pay attention prior to applying the approximation. As mentioned earlier, the expansion in \eqref{Cheb_approx} is valid only for $x \in [-1, 1]$. In the present case, the target function is $h(x) = \exp(x)$, where $x= \tau t/ \sigma_r^2$. Recall that $t = f(i-j)$ and $\tau = f(i)$, which take values in the intensity range $[L, U]$.
Since the Chebyshev polynomials are defined on a symmetric domain, we first apply the following centering:
\begin{equation}
\label{affine}
g(i) =  f(i) - t_c  \qquad (i \in I),
\end{equation}
where $t_c=(L+U)/2$. Let $\mu = (U-L)^2/4\sigma_r^2$. 
One can verify that $g(i) g(i-j)/\sigma_r^2 \in [-\mu, \mu]$ for all $i \in I$ and $j \in \Omega$.
Thus, we are required to approximate $h(x)$ over the interval $[-\mu, \mu]$. To do so, we consider the following rescaled version of \eqref{Cheb_approx}:
\begin{equation}
\label{Cheb_approx_tau}
 h(x) \approx \sum_{l = 0}^{N} d_l T_l(x/\mu).
\end{equation}
For this expansion, formula \eqref{coeff} becomes
\begin{equation*}
 d_l =  \begin{dcases}
 \frac{1}{N+1} \sum_{k=1}^{N+1} \! h(\mu \xi_k), \ \  & l = 0\\
 \frac{2}{N+1} \sum_{k=1}^{N+1} \! h(\mu \xi_k) T_l(\xi_k),  \ \ & l \neq 0.
 \end{dcases}
\end{equation*}
Plugging \eqref{expansion} into \eqref{Cheb_approx_tau}, we obtain  
\begin{equation}
\label{Cheby}
h(x) \approx   \sum_{n = 0}^{N} c_n x^n, \quad  c_n =  (1/ \mu)^n \sum_{l=n}^{N}  d_l a_{l,n}.
\end{equation}

In summary, the sequence of operations for computing \eqref{approxBF} using the Gauss-Chebyshev approximation, which we will henceforth refer to as the Gauss-Chebyshev Filter (GCF), is as follows:
\begin{itemize}
\item Center the image using \eqref{affine}.
\item Compute $\widehat{\mathcal{B}[g]}$ using the Chebyshev approximation in \eqref{Cheby}. This is done using the fast algorithm in section \ref{sec:FBF}. 
\item Undo the centering: $\widehat{\mathcal{B}[f]}  = \widehat{\mathcal{B}[g]} + t_c$.
\end{itemize}

\section{Simulations}
\label{sec:SIM}

We first present a couple of representative results  in Figures \ref{fig:tau100} and \ref{fig:vary_tau} to demonstrate that the Gauss-Chebyshev approximation is indeed better that the Gauss-Polynomial approximation. Indeed, the Gauss-Chebyshev approximation of the translated kernels are more accurate, particularly for large $\tau$.

\begin{table}[!htp]
\setlength{\tabcolsep}{3.5pt}
\caption{MSE as a function of the degree $N$ when $\sigma_s = 5$ and $\sigma_r = 30$. The Checker image \cite{SIPI} was used for the experiment.}
\vspace{2mm}
\centering 
\begin{tabular}{| c | c  c  c  c  c  c c |}
\hline
{Method}$\backslash${$N$}   
&$4$  &$ 8$ & $10$  &$12$   &$16$     &$20$    &$25$  
\\ \hline
GPF & 33.54  & 30.45 & 28.63 & 26.57  & 21.39  & 14.05  & 0.24  \\
\hline
GCF & {\bf 7.14}  & {\bf 4.23} & {\bf -11.42}  & {\bf -23.37}  & {\bf -39.88}  & {\bf -40.54} & {\bf -40.54} 
\\
\hline			
\end{tabular}
\label{tab:ckb}
\end{table}

\begin{figure}
\centering
\subfloat[Checker ($150 \times 150$) \cite{SIPI}.]{\includegraphics[width=0.40\linewidth]{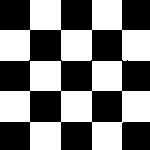}} \hspace{0.1 cm}
\subfloat[Bilateral Filter, \textbf{45} sec.]{\includegraphics[width = 0.40\linewidth]{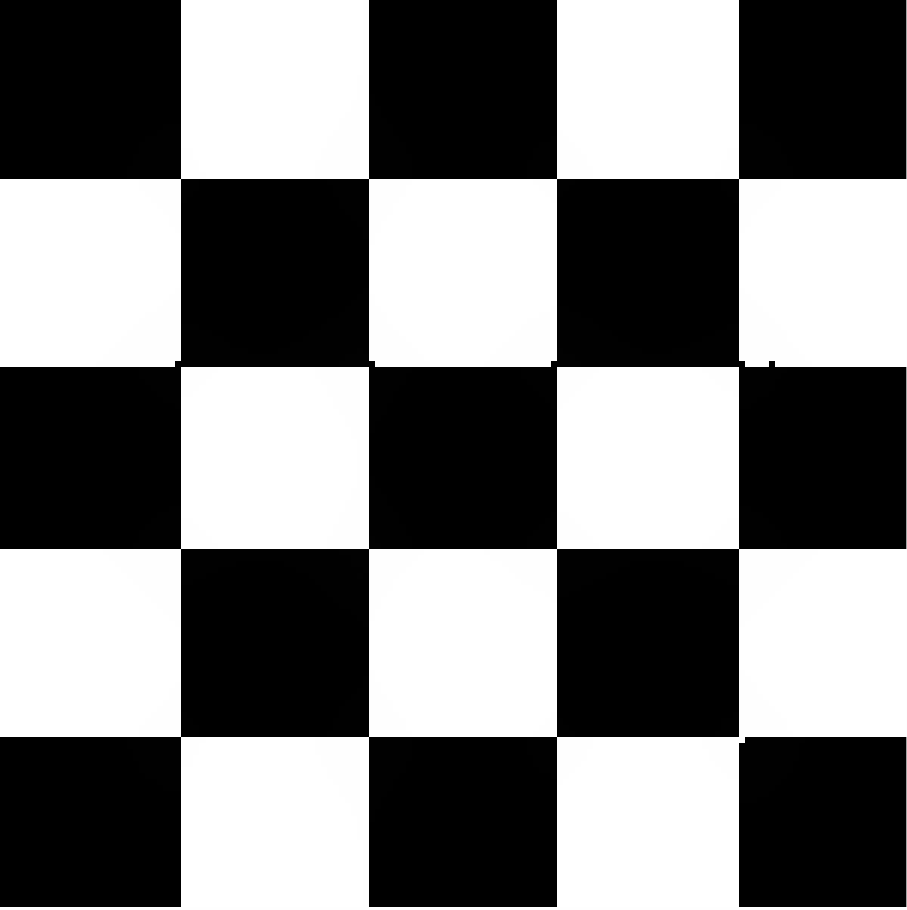}} \\
\subfloat[GPF \cite{Chaudhury2015}, \textbf{28.63} dB, \textbf{31 ms}.]{\includegraphics[width=0.40\linewidth]{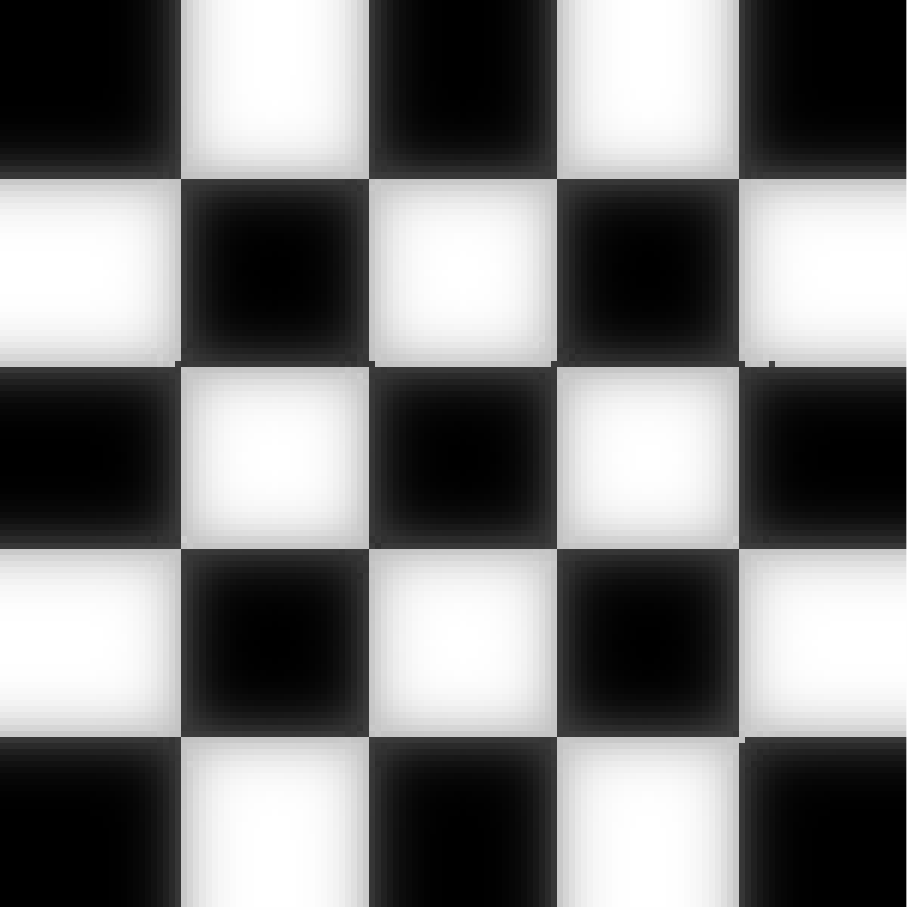}} \hspace{0.1 cm}
\subfloat[GCF, \textbf{-11.42} dB, \textbf{32 ms}.]{\includegraphics[width=0.40\linewidth]{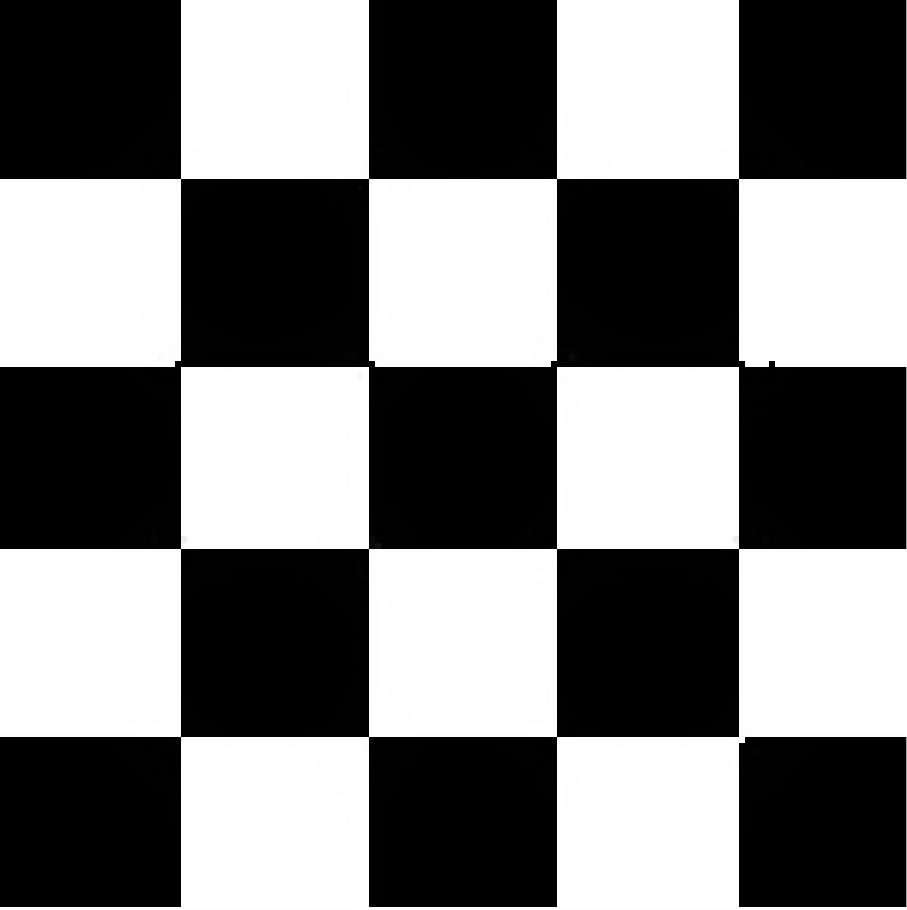}} 
\caption{Comparison of the direct implementation of \eqref{BF} with GPF \cite{Chaudhury2015} and the proposed GCF algorithm. The parameters of the filter are: $\sigma_s =5$ and $\sigma_r  = 30$. The degree of the Taylor and Chebyshev approximation is $N = 10$. The MSE and the corresponding run-times are shown in bold. Notice the visible blurring in (c) due to the poor approximation.
} 
\label{ckb}
\end{figure}

We next present results concerning the accuracy and the run-time. The simulations were performed using Matlab on a $3.4$ GHz Intel $8$-core machine with $32$ GB memory. To quantify accuracy, we have used the mean-squared-error (MSE) between \eqref{BF} and the filtering obtained using the corresponding fast algorithm.
The MSE between two images $f : I \rightarrow \mathbb{R}$ and $g : I \rightarrow \mathbb{R}$ is defined to be $10 \log_{10}(\text{MSE})$ dB, where $\text{MSE} = |I|^{-1}\sum_{i \in I}(f(i)-g(i))^2$.
For the simulation, we have used the Checker and Testpat images from \cite{SIPI}, and the popular Barbara image. 
The former images have lot of sharp edges, while the latter has a fair mix of texture and edges.
The accuracies offered by the Gauss-Polynomial and the Gauss-Chebyshev Filter are compared in Table \ref{tab:ckb}.
It is not surprising that the latter performs significantly better. This is also evident from the visual results in Figure \ref{ckb}. 

\begin{table}
\setlength{\tabcolsep}{3.5pt}
\caption{Comparison of run-time (sec) / MSE of the proposed GCF algorithm and the fast algorithms in \cite{Paris2006, Yang2009, Chaudhury2013, Chaudhury2015}. We perform the comparison at different $\sigma_s$, using Testpat  \cite{SIPI} as the test image. We fixed $\sigma_r = 30$ and $N = 28$. For the exact implementation, we just report the run-times (sec).} 
\vspace{2mm}
\centering 
\begin{tabular}{| c | c | c | c | c | c | c|}
\hline
$\sigma_s$   &Exact   &\cite{Paris2006}   &\cite{Yang2009}   &\cite{Chaudhury2013}  &\cite{Chaudhury2015} & GCF \\ [+0.5ex]
\hline  
2 & 103  & 0.67 / -3.0    & 1.01 / 6.0  & 1.18 / -6.3   & 0.46 / -8.8 & 0.47 / \bf{-40.7} 
\\ [+0.5ex]
3 & 238  & 1.13 / -4.3    & 1.05 / 7.0  & 1.24 / -4.6   & 0.58 / -6.6  & 0.59 / \bf{-38.9} 
\\ [+0.5ex]
4 & 370   & 1.27 / -2.4    & 1.21 / 7.8  & 1.35 / -1.1  & 0.65 / -5.8  & 0.64 / \bf{-37.4}
\\ [+0.5ex]
5 & 570  & 1.73 / -2.3    & 1.54 / 8.3  & 1.46 / 2.5  & 0.90 / -5.6  & 0.89 / \bf{-36.3}
\\ [+0.5ex]
10 & 2250  & 2.31 / 0.7   & 1.64 / 10.2  & 1.92 / 0.2  & 1.04 / -6.1  & 1.01 / \bf{-32.2}
\\ [+0.5ex]
15 & 5207  & 11.2 / 2.5    & 2.00 / 11.4  & 2.30 / 1.0  & 1.83 / -6.7 & 1.72 / \bf{-20.4} \\[+0.5ex]
\hline			
\end{tabular}
\label{tab:TP}
\end{table}

\begin{figure}
\centering
\subfloat[Testpat ($1024 \times 1024$) \cite{SIPI}.]{\includegraphics[width=0.47\linewidth]{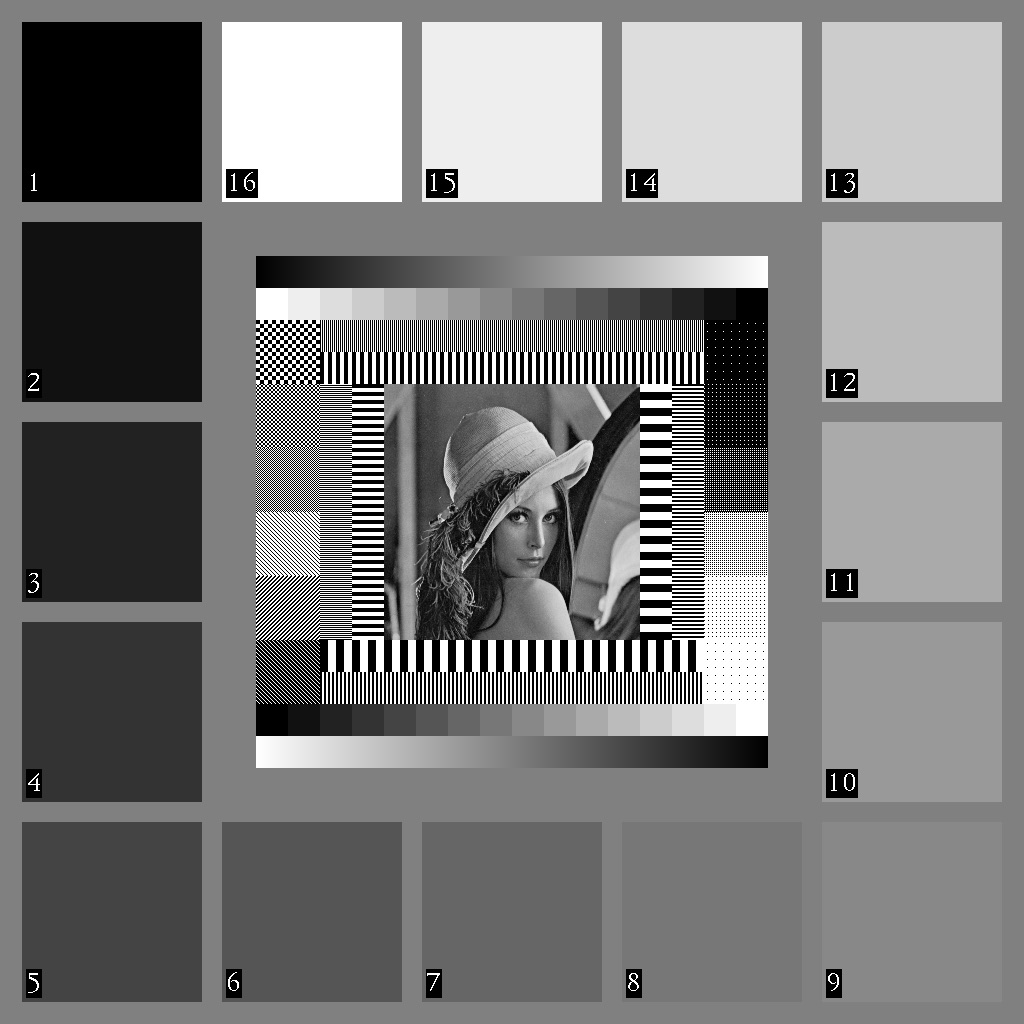}}  \hspace{0.01 cm}
\subfloat[Bilateral Filter.]{\includegraphics[width=0.47\linewidth]{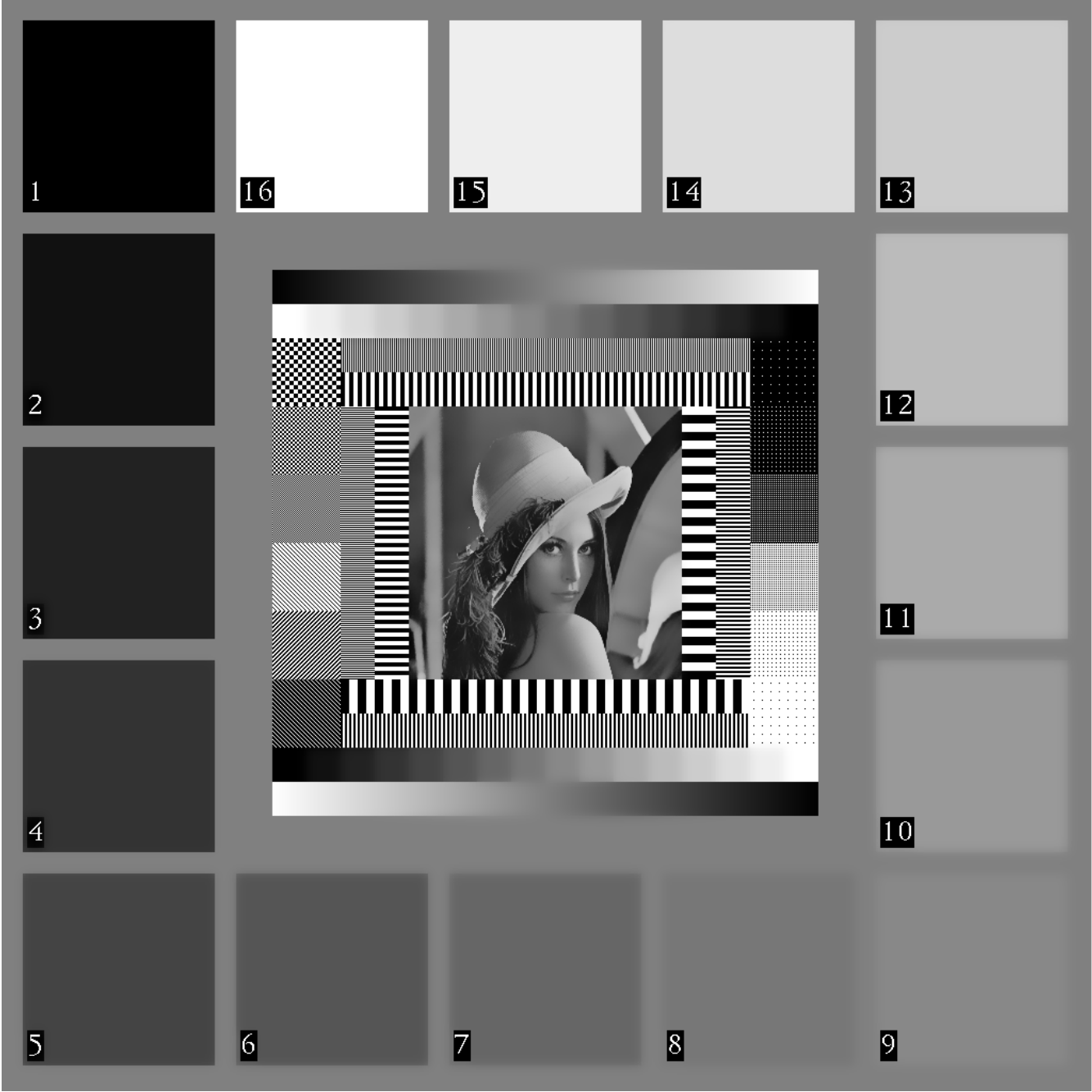}} \\ 
\subfloat[Bilateral Grid \cite{Paris2006}, \textbf{-2.25} dB.]{\includegraphics[width=0.47\linewidth]{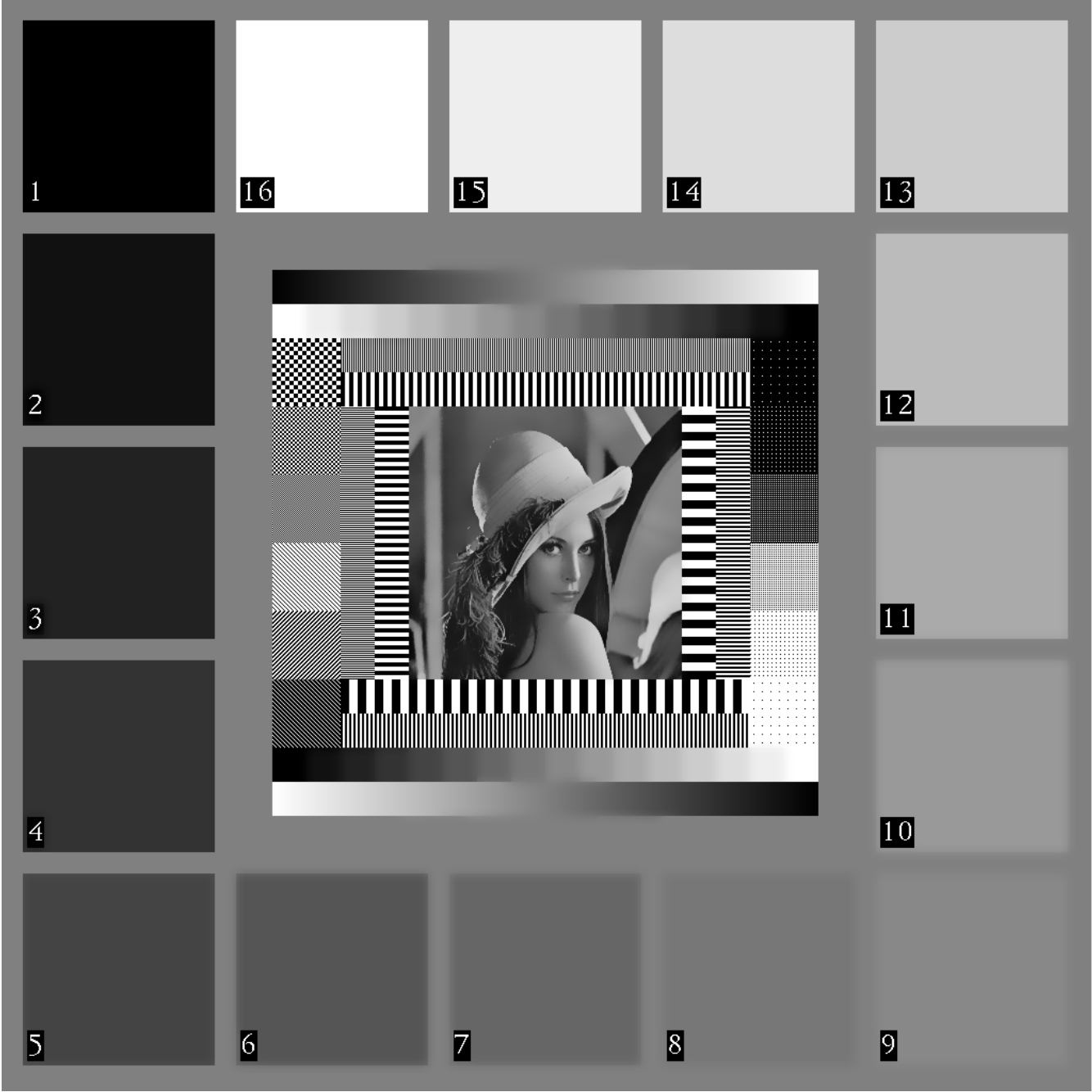}} \hspace{0.01 cm}
\subfloat[Range Interpolation \cite{Yang2009}, \textbf{8.34} dB.]{\includegraphics[width=0.47\linewidth]{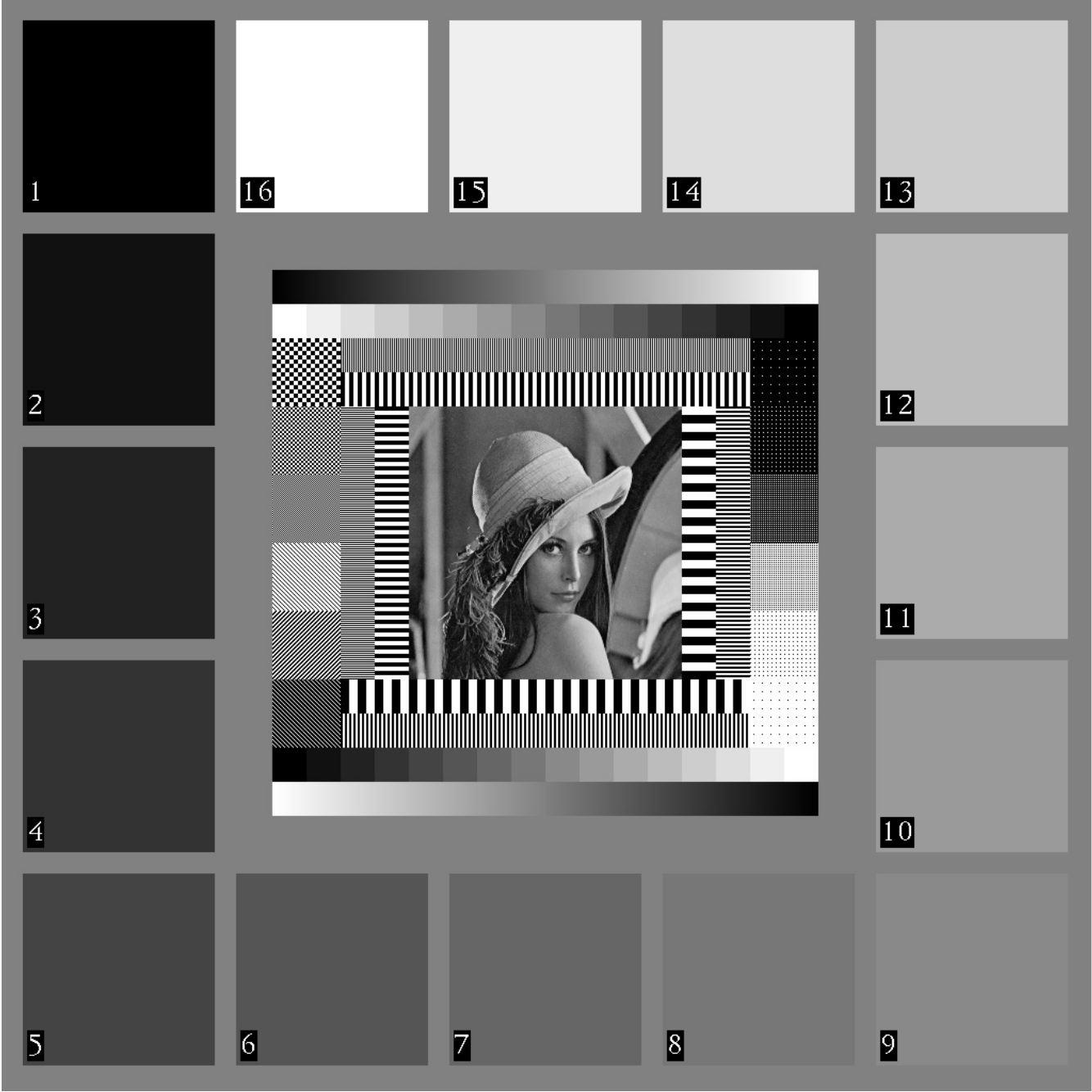}} \\
\subfloat[GPF \cite{Chaudhury2015}, \textbf{1.22} dB.]{\includegraphics[width=0.47\linewidth]{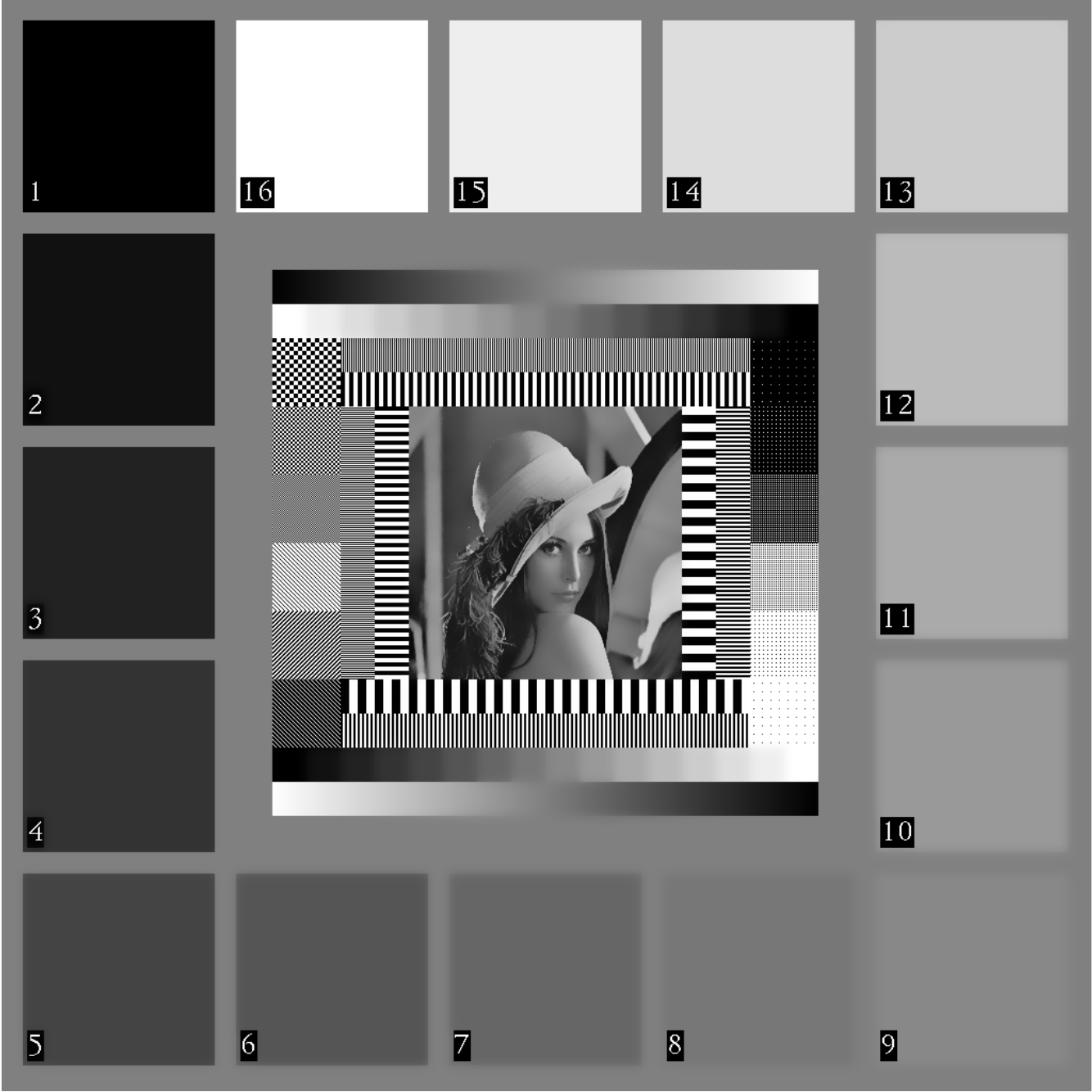}} \hspace{0.01 cm}
\subfloat[Proposed GCF, \textbf{-16.35} dB.]{\includegraphics[width=0.47\linewidth]{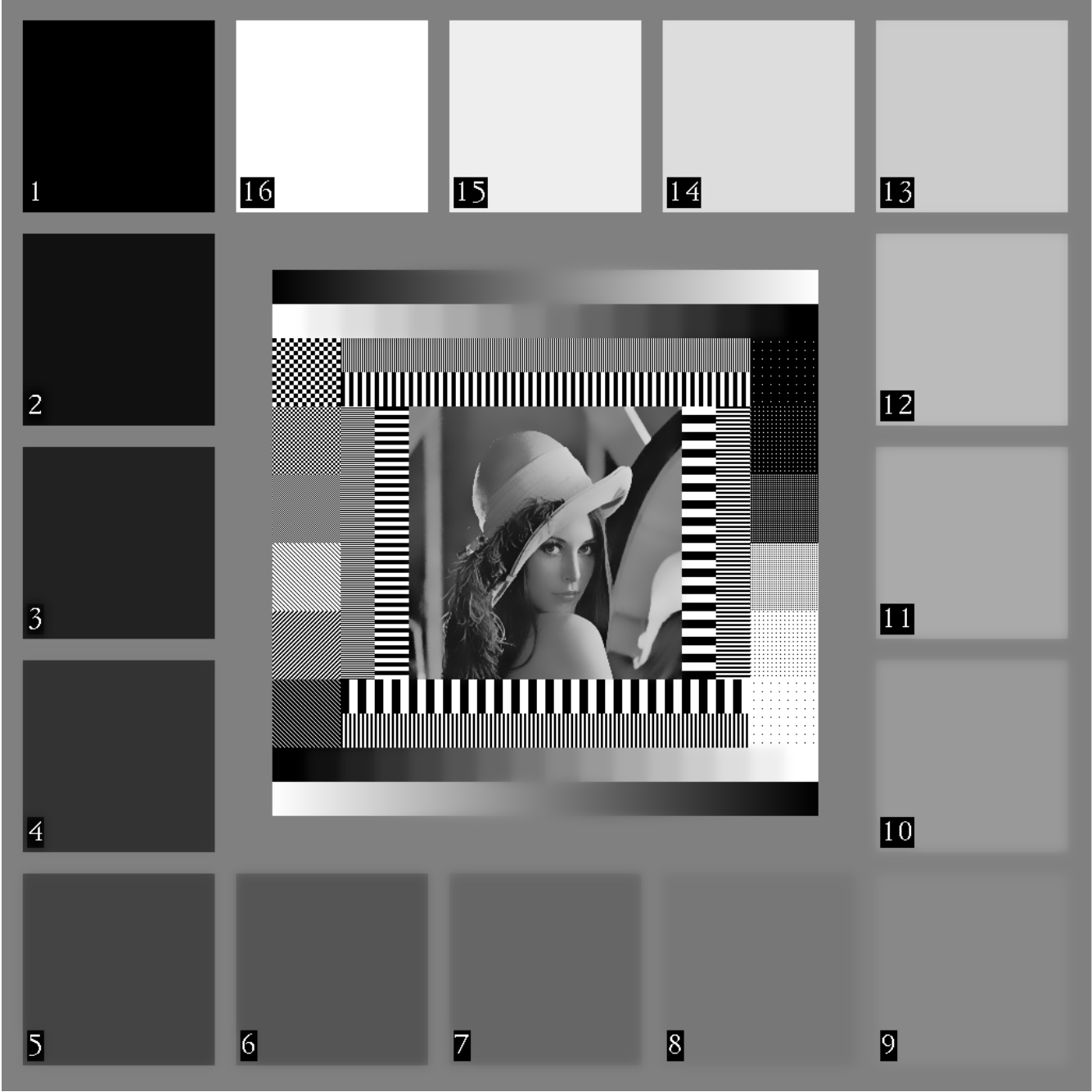}} 
\caption{Comparison of the direct implementation with various fast algorithms on a montage. The filter parameters are $\sigma_s =5$ and $\sigma_r  = 30$. For GPF and GCF, the approximation degree is $N=26$.} 
\label{testpat}
\end{figure}

\begin{figure}
\centering
\subfloat[Barbara ($512 \times 512$).]{\includegraphics[width=0.44\linewidth]{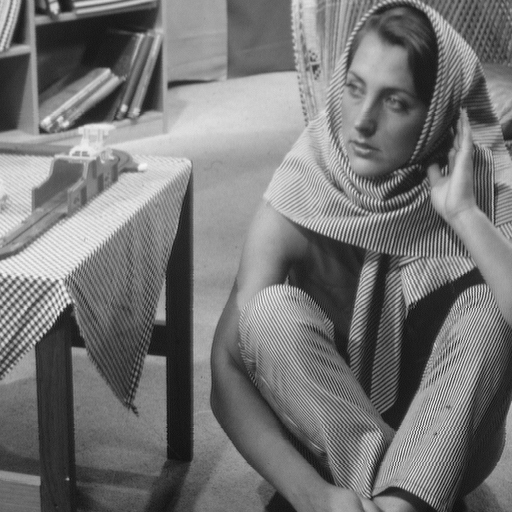}} \hspace{0.01 cm}
\subfloat[Bilateral Filter, \textbf{125} sec.]{\includegraphics[width = 0.44\linewidth]{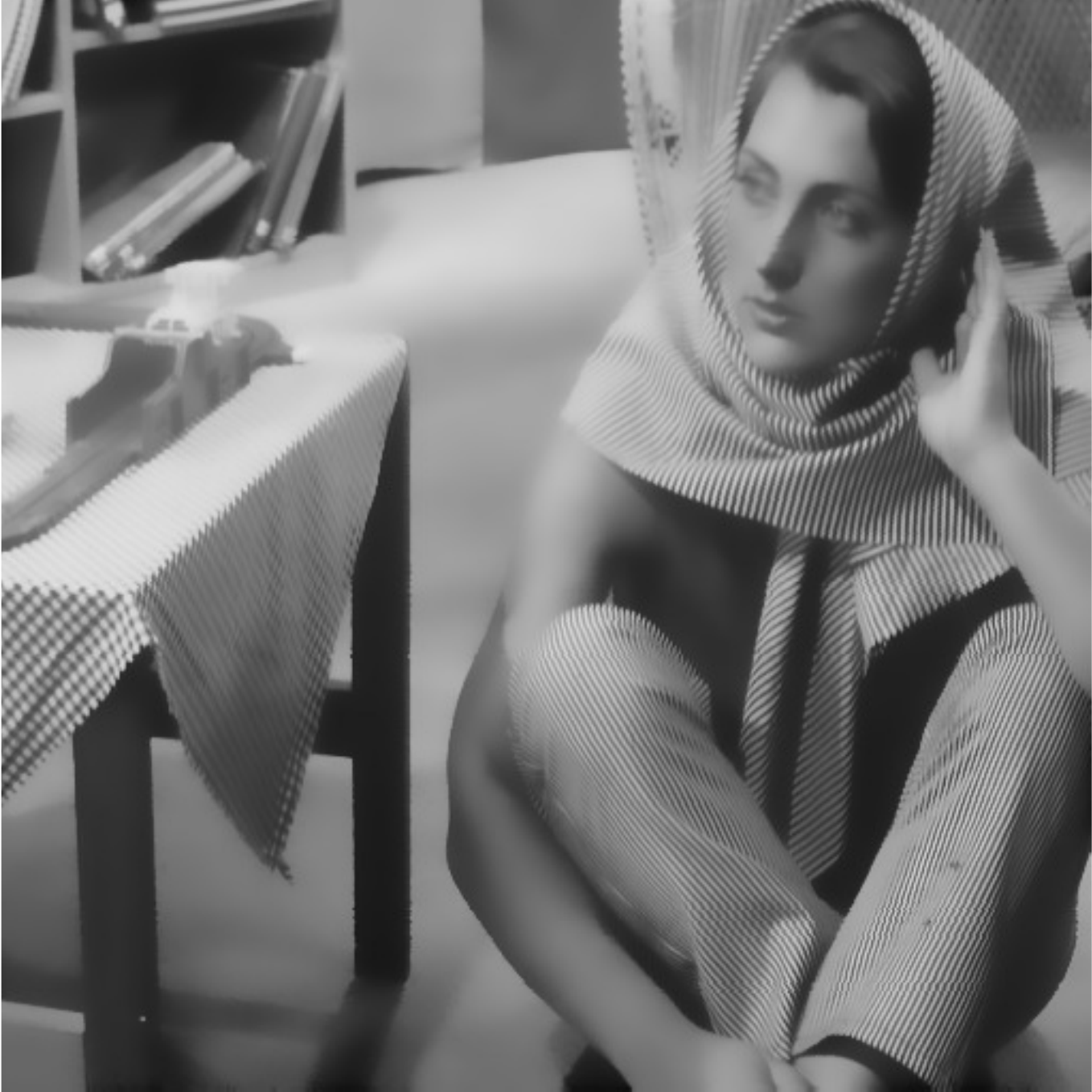}} \\
\subfloat[GPF, \textbf{-2.13} dB, \textbf{154 ms}.]{\includegraphics[width=0.44\linewidth]{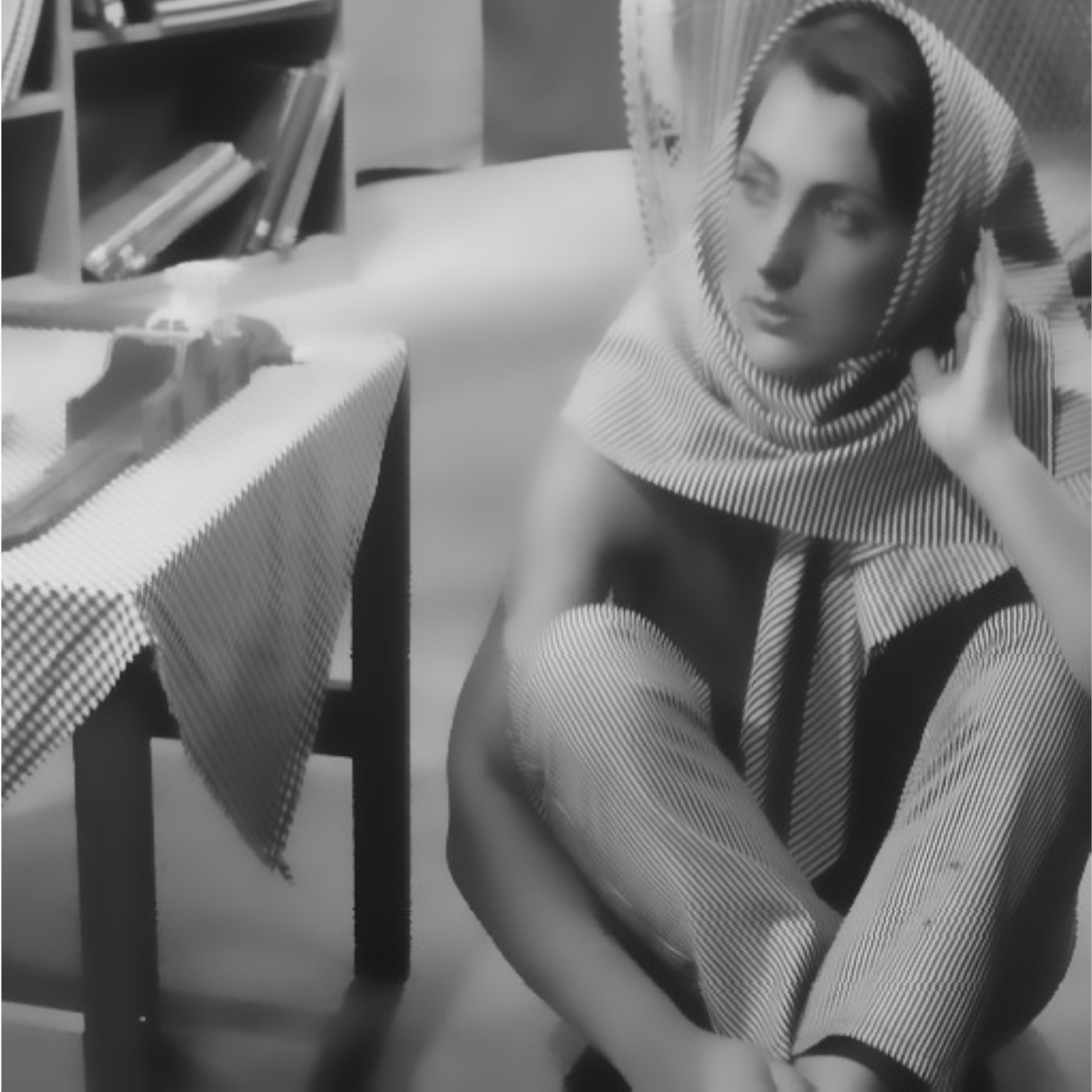}} 
\hspace{0.01 cm}
\subfloat[GCF, \textbf{-2.13} dB, \textbf{156 ms}.]{\includegraphics[width=0.44\linewidth]{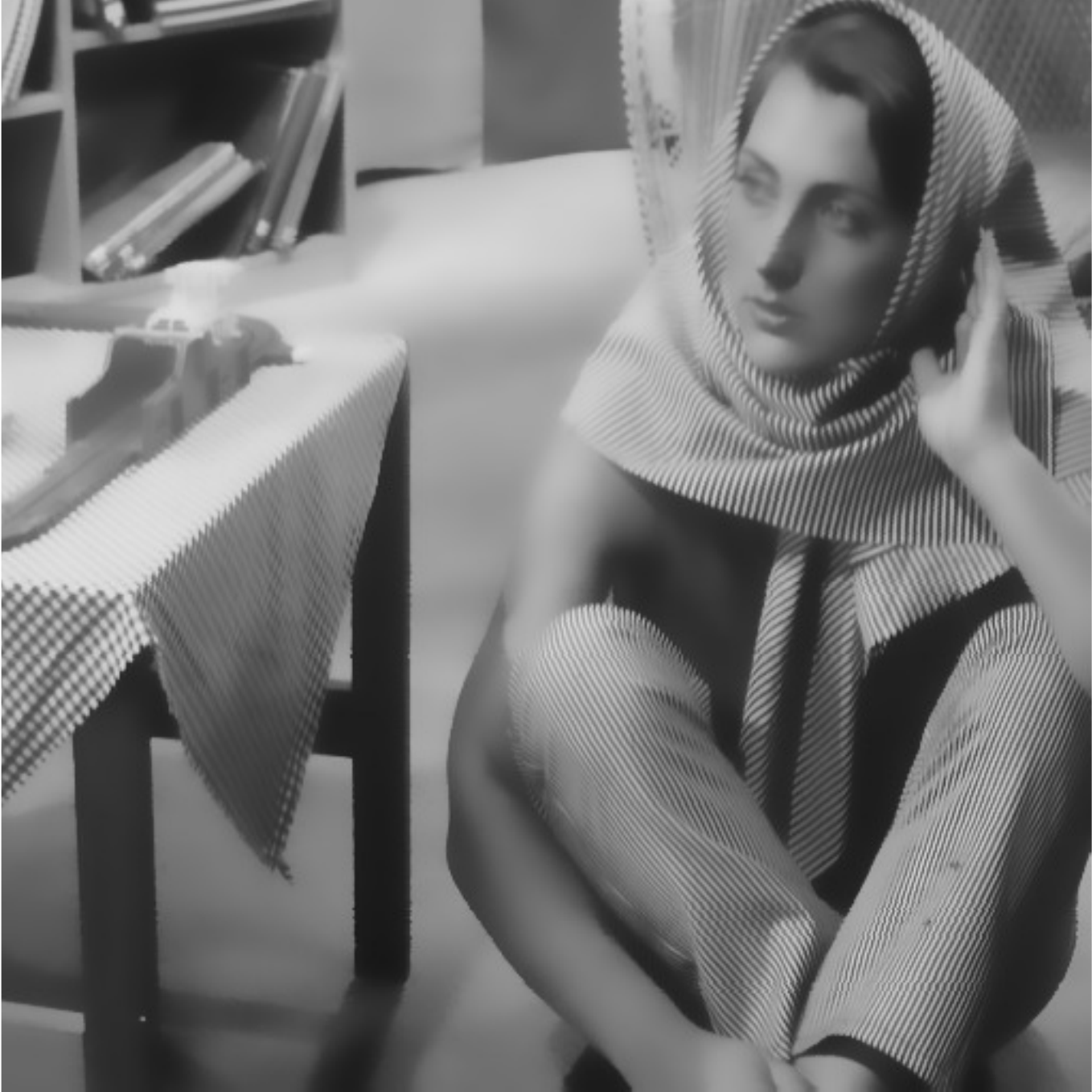}} 
\caption{Comparison of the direct implementation of \eqref{BF} with GPF \cite{Chaudhury2015} and the proposed GCF algorithm. The parameters of the filter are: $\sigma_s =5$ and $\sigma_r  = 30$. The degree of the polynomial is $N = 30$.} 
\label{barbara}
\end{figure}

Further visual results are provided in Figures  \ref{testpat} and \ref{barbara}. Notice in Figure \ref{testpat} that the GCF filtering is significantly better than some of the existing fast algorithms.
On the other hand, GPF and GCF have almost identical performance for the Barbara image in Figure \ref{barbara}. This is not surprising given that this image has far less sharp edges compared to the other images.

Finally, we compare the run times and the accuracy of various fast algorithms and report the results in Table \ref{tab:TP}. 
We see that the accuracy of GCF is substantially better that that offered by some of the existing fast algorithms, while its run-time is comparable. In general, notice that the run-time of GCF is few orders smaller than that of the direct implementation of \eqref{BF}.

\section{Conclusion}
\label{sec:CONC}

We proposed a fast algorithm for Gaussian bilateral filtering based on the Gauss-Chebyshev approximation. In particular, we demonstrated that the algorithm is comparable with the GPF algorithm \cite{Chaudhury2015} in terms of run-time, but performs significantly better on images that have lot of sharp edges. We note that the GPF algorithm was already shown in \cite{Chaudhury2015} to be superior to existing fast algorithms for natural images, such as the Barbara image in Figure \ref{barbara}.
The proposed algorithm can be particularly useful for regularizing depth maps arising in stereo reconstruction, which exhibits sharp transitions.
For example, it was demonstrated in \cite{YK2006} that a local aggregation of the similarity score using bilateral weights significantly improves the 
disparity estimation. 

\bibliographystyle{IEEEbib}

\end{document}